# Geometry-Based Multiple Camera Head Detection in Dense Crowds

Nicola Pellicanò
nicola.pellicano@u-psud.fr

Emanuel Aldea
emanuel.aldea@u-psud.fr

Sylvie Le Hégarat-Mascle
sylvie.le-hegarat@u-psud.fr

SATIE Laboratory
CNRS UMR 8029
Paris-Sud University
Paris-Saclay University
Orsay 91405, France

## Abstract

This paper addresses the problem of head detection in crowded environments. Our detection is based entirely on the geometric consistency across cameras with overlapping fields of view, and no additional learning process is required. We propose a fully unsupervised method for inferring scene and camera geometry, in contrast to existing algorithms which require specific calibration procedures. Moreover, we avoid relying on the presence of body parts other than heads or on background subtraction, which have limited effectiveness under heavy clutter. We cast the head detection problem as a stereo MRF-based optimization of a dense pedestrian height map, and we introduce a constraint which aligns the height gradient according to the vertical vanishing point direction. We validate the method in an outdoor setting with varying pedestrian density levels. With only three views, our approach is able to detect simultaneously tens of heavily occluded pedestrians across a large, homogeneous area.

# 1 Introduction

Pedestrian detection is a fundamental task in computer vision, closely related to applications such as video surveillance, autonomous driving or action recognition. Some characteristics of the analyzed scene and camera setup improve significantly the reliability of the detection: a low pedestrian density, close to vertical optical axis, and a good resolution representation of individuals. However, the recent focus on the analysis of large, densely crowded outdoor areas underlines the current limitations in presence of persistent, heavy clutter.

Detection strategies based on multiple overlapping views may be used to achieve more robust inference provided that *sensor data are fused* prior to detection. This still leaves the problem of *how to associate data* among views which may exhibit significant geometric and photometric variation. Above all, the *joint projection* of visual information in a common reference system is conditioned by an accurate estimation of the relative camera poses, and of the ground plane. As some of the underlying assumptions are violated (i.e. persistent clutter for foreground extraction, heavy occlusions for part-based detectors, homogeneous crowd dynamics for independent motion based inference), the detector breaks down. Moreover, resilience to camera pose variations or to people appearance comes at a cost, in the form of human intervention for calibration procedures or for scene-dependent supervised learning.

There exists a large body of research for pedestrian detection, therefore the following paragraphs focus on approaches suited for moderate to high density scenes. For identifying strongly occluding people, multiple cameras are better positioned in order to resolve ambiguities in at least a subset of the available views. Depending on how sensor data are combined, detection methods rely on raw data level (low-level), on object level (mid-level) or on trajectory level (high-level) data fusion.

**Scene geometry** Difficult detection scenarios benefit from strategies which avoid performing the detections in individual views. However, the lower the fusion level, the greater will be the impact of the geometry alignment among the cameras and the scene. In a number of studies, the relative camera poses and the ground plane orientation are identified by relying on a robust estimation of the ground plane homography using ground inliers [13, 23]. A precise estimation requires a large uniformly distributed set of observations which are difficult to obtain, considering the homogeneity of the ground and the potentially significant pose variations among cameras. Some studies rely on manual ground annotations [5, 11], but for such a solution to be constrained accurately across the work area, the annotations should ideally be uniformly and densely performed. Finally, classical extrinsic calibration relies on specific objects being observed in multiple views prior to the analysis, but this approach does not scale to large outdoor areas. Moreover, the concomitant presence of calibration objects in different views [6] during analysis is non-viable in cluttered realistic conditions.

**Ground plane projection** A common ground plane hypothesis is adopted by most multiple view based detectors (a notable exception being [1]), due to the simplificatory assumptions it allows for data fusion. In order to simplify data association among cameras while at the same time avoiding the difficult detection task in single views, the vast majority of subsequent works rely on foreground extraction and the combination of foreground maps in the ground plane reference, as opposed to high-level approaches which apply multi-view homographies onto single-view detections [14]. Early works such as [18] relied on basic appearance cues such as color in order to find correspondences across cameras. In subsequent studies, the data fusion may be performed under various forms, such as a probabilistic occupancy map relying on a generative model [7], silhouette based extraction [2, 10], stochastic spatial models [8, 26], or a joint foreground and appearance likelihood objective function [17].

**Multiple homography methods** Methods such as the ones proposed by Khan and Shah [13] or by Eshel and Moses [6] rely on a detection performed at varying heights with respect to the reference plane. However, since the objective of [13] is to locate objects at ground level, the method is sensitive to strong clutter. Conversely, the method proposed in [6] copes better with strong occlusions, as it compares intensity values of different views at the same projected location to locate heads. However, an intensity-correlation based matching can be highly sensitive to illumination variations, typical to outdoor settings, and to large perspective changes, typical to wide baseline camera setups. Moreover, while Khan and Shah [13] rely on non metric multiple homographies, Eshel and Moses [6] estimate the metric scale by manual intervention.

All the methods above, other variations based on 3D carving [21, 22] and extensions based on the spatio-temporal evolution of detections [3, 12, 20] rely heavily at an incipient stage on foreground extraction, and are significantly impaired if the foreground cannot be segmented. A typical example is [13] which requires feet visibility in order to work. Unfortunately, cameras placed with low pitch angles as it is generally the case for high-density crowd surveillance would not observe sufficient empty areas among proximate pedestrians

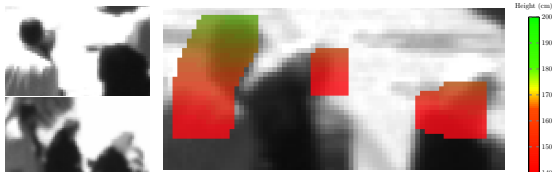

Figure 1: Head detection detail with a height map overlay in the central camera view. Two additional views (see left) are used for the height map estimation. Note the height gradient following the local vertical direction, and the middle detection for which a strong occlusion is present in one of the lateral views.

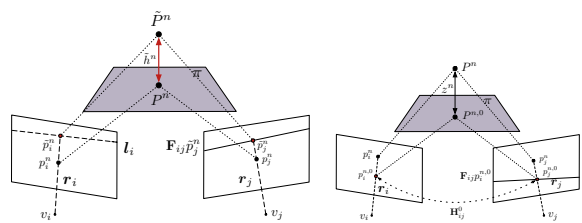

(a) computing a height hypothesis $\tilde{h}^n$

(b) ground projection $(p_i^{n,0}, p_j^{n,0})$ from inlier pair $(p_i^n, p_j^n)$.

Figure 2: Ground identification geometry.

in order to benefit from foreground extraction.

We are inspired by the work of Eshel and Moses [6], who underline the graceful degradation of homography-based head detection as crowd density increases. Our contributions are summarized as follows: **(i)** we demonstrate a new fully unsupervised method for relative camera pose and homography estimation which avoids placing calibrating objects inside the investigated area, either during the detection algorithm or prior to the analysis **(ii)** we rely on a data association cost among camera views which is able to cope with intensity and perspective variations specific to outdoor, wide baseline configurations **(iii)** we express the head detection problem as a stereo MRF-based optimization of a dense pedestrian height map. By addressing these key points, we believe the proposed advances will improve the applicability of geometry-based strategies for head detection to large, cluttered scene analysis (Fig. 1).

## 2 Inferring scene and camera geometry

The acquisition system consists of a central reference camera $C_i$ and a set of neighboring cameras $\mathcal{N}(C_i)$. The geometry analysis can be divided into two parts.

First, the epipolar geometry between pairs of adjacent cameras needs to be estimated, and the relative camera poses extracted. Then, following the idea of Eshel and Moses [6], we restrict the search space to a volume contained between two planes parallel to the ground plane. This amounts to detecting pedestrians with heights in a specified interval $[h_{min}, h_{max}]$. In terms of camera geometry, this requires a metric registration of each camera with respect to the ground plane in terms of variable-height homographies.

**Relative pose estimation** A fundamental matrix $\mathbf{F}_{ij}$ between $C_i$ and each $C_j \in \mathcal{N}(C_i)$ is estimated using the unsupervised method of Pellicanò et al. [19], by accumulating inliers $S_{ij}$ from a pair of synchronized video streams of a crowded area recorded using normal or wide baselines. From each $\mathbf{F}_{ij}$, the relative pose $(\mathbf{R}_{ij}, \mathbf{t}_{ij})$ is obtained by SVD decomposition. At this point, we also enforce a metric scale $t_{ij}^m$ for each pair by setting the norm of $\mathbf{t}_{ij}$ to the actual distance $D_{ij}^l$ between the cameras measured with a standard handheld laser device: $t_{ij}^m = D_{ij}^l \cdot \mathbf{t}_{ij}/\|\mathbf{t}_{ij}\|$. This simple operation is *the only manual procedure* we require in order to inject the real-world scale of the scene into the estimations.

The following mandatory step is to enforce a common metric scale to all the camera poses, as any imprecision introduced in the computation of the different $(\mathbf{R}_{ij}, \mathbf{t}_{ij})$ will have a negative impact on the data association. For any triplet $C_{i-j-k}$, we rely on triple matches (which are a subset of all matches identified during the fundamental matrix estimation) to propagate the scale, while including the simple matches as well in the BA in order to stabilize the other degrees of freedom of the problem. Then, if more then three cameras are used, a global BA may be applied over all the poses and available observations.

**Variable-height homographies** Let us consider the estimation of the variable-height homography between the reference camera $C_i$ and some neighbor camera $C_j$. The following steps are to applied as well to any pair $(C_{i-j})$, with $C_j \in \mathcal{N}(C_i)$. The variable-height homography $\mathbf{H}_{ij}^h$ can be derived from Criminisi *et al.* [4] as $\mathbf{H}_{ij}^h = \mathbf{B}_j^h \mathbf{H}_{ij}^0 (\mathbf{B}_i^h)^{-1}$, where $\mathbf{H}_{ij}^0$ is the homography constraint between $C_i$ and $C_j$ on the ground plane ($h = 0$), and $\mathbf{B}_k^h$ represents the homology at a specific camera $C_k$ and height $h$. The homology transforms, expressed in the image reference system, link a ground location to their corresponding projection on the plane parallel to the ground at height $h$, and are expressed as $\mathbf{B}_k^h = \mathbf{I} + \alpha_k h \mathbf{v}_k \mathbf{l}_k^T$, where $\mathbf{I}$ is the $3 \times 3$ identity matrix, $\alpha_k$ is the metric scale coefficient, $\mathbf{v}_k$ is the vertical vanishing point and $\mathbf{l}_k$ is the unit-length vanishing line of the reference ground plane.

The $\mathbf{v}_k$ and $\mathbf{l}_k$ for each camera are estimated under Manhattan world assumptions by using the method of Lezama *et al.* [16]. The underlying justification is that, although urban repetitive patterns are difficult to match reliably for inferring relative poses, they can be used in *individual* views for estimating $\mathbf{v}_k$ and $\mathbf{l}_k$.

The estimation of the ground plane homography $\mathbf{H}_{ij}^0$ can be carried out by detecting in the two images point matches lying on the desired plane. In order to perform an automated estimation of $\mathbf{H}_{ij}^0$, we propose to extract a candidate set of point matches from the inlier set $S_{ij}$ provided by the computation of $\mathbf{F}_{ij}$. The point extraction from video relies on the dynamics of people moving in the scene for the registration [19]. This implies that the final 3D cloud of inliers can be clustered into points belonging to dominant planes (ground and building facades), and into points originated from pedestrian bodies moving across the scene. Given a point correspondence $(p_i^n, p_j^n)$ in the two views, we assign to it a label $\tilde{h}^n$, corresponding to the estimated camera height under the assumption that the related 3D point is on the ground, which can be easily calculated as follows by using the epipolar geometry and the vanishing points. Only point matches located under the camera vanishing lines are considered.

Let us call $\mathbf{r}_i^n$ the straight line passing through $p_i^n$ and $\mathbf{v}_i$ (we define $\mathbf{r}_j^n$ accordingly). As depicted in Fig. 2(a), the position of a point $\tilde{p}_i^n$ is evaluated as the intersection between $\mathbf{r}_i^n$ and the vanishing line. The point, $\tilde{p}_i^n$ corresponds to the projection of $p_i^n$ on a plane which is parallel to the ground and at the height of the camera. In the second view, the point $\tilde{p}_j^n$ is detected as the intersection between the epipolar line $\mathbf{F}_{ij}\tilde{p}_i^n$ and $\mathbf{r}_j^n$. Let us call $P^n$ and $\tilde{P}^n$ the 3D points obtained from the triangulation of $(p_i^n, p_j^n)$ and $(\tilde{p}_i^n, \tilde{p}_j^n)$ respectively. Then, $\tilde{h}^n = \| P^n - \tilde{P}^n \|$.

We turn then the ground identification step into a data clustering problem with random variable $\tilde{H}^n$. We intend to separate the set of points generated by moving pedestrians on the ground (occasional visible feet locations, shadows) from the points generated by body parts in the presence of other matches (i.e. building interest points), that we will define as *outliers* of the process. We perform this robust estimation task using the method of Gebru *et al.* [9], which introduces an EM-algorithm robust to data outliers.

Once $\mathbf{H}_{ij}^0$ is estimated, the metric scale coefficients $\alpha_i$ and $\alpha_j$ may be computed. According to Criminisi *et al.* [4], $\alpha_k$ (with $k = \{i, j\}$) can be derived from an image point, along with its projection on the ground plane, and with their metric distance in the real world. A convenient way of computing $\alpha_k$ is to take two reference points at a known distance on the scene. Our procedure for the automatic estimation of the $\alpha_k$ values consists in exploiting the set of inlier matches $S_{ij}$, in such a way that every pair of the set, which does not corre-spond to a point on the ground plane, votes for global candidates $\alpha_i$ and $\alpha_j$. Given the match $(p_i^n, p_j^n)$, the calculation of $\alpha_k$ requires the identification of a corresponding match of their projection on the ground $(p_i^{n,0}, p_j^{n,0})$ (see Fig. 2(b)). Given the ground plane homography $\mathbf{H}_{ij}^0$, the identification can be cast as the following optimization problem:

$$\arg\min_{p_i^{n,0}, p_j^{n,0}} \left[ d\left(\mathbf{H}_{ij}^0 p_i^{n,0}, p_j^{n,0}\right)^2 + d\left((\mathbf{H}_{ij}^0)^{-1} p_j^{n,0}, p_i^{n,0}\right)^2 \right] \quad s.t. \quad p_i^{n,0} \in \mathbf{r}_i, p_j^{n,0} = \mathbf{F}_{ij} p_i^{n,0} \cap \mathbf{r}_j \quad (1)$$

where $d(\cdot,\cdot)$ is the Euclidian distance operator, and $\cap$ indicates the intersection between two lines. With the given constraints the problem can be reduced to a single variable optimization problem, as the cost can be expressed as a function of one of the two components $p_i^{n,0} = (x_i^{n,0}, y_i^{n,0})$. We obtain a solution with the Levenberg-Marquardt algorithm, by choosing as starting value $y_0 = y_i^n$ of $p_i^n = (x_i^n, y_i^n)$. Then, $\alpha_i$ and $\alpha_j$ may be obtained from the point pairs $(p_i^n, p_i^{n,0})$ and $(p_j^n, p_j^{n,0})$ respectively.

Once every match within $S_{ij}$ votes for a hypothesis $(\alpha_i, \alpha_j)$, we select the variable-height homography model which best fits the given data. We then define a reprojection error corresponding to a given derivation of the $\mathbf{H}_{ij}^h$; finally, an LMedS estimator is applied to select the best model.

For a generic pixel $p_i$ in the image space of camera $C_i$, we define $p_{j,min} \sim \mathbf{H}_{ij}^{h_{min}} p_i$ and $p_{j,max} \sim \mathbf{H}_{ij}^{h_{max}} p_i$. These two points lie on the epipolar line $\mathbf{F}_{ij} p_i$ in the image space of camera $C_j$, and define the extremes of an *epipolar segment* which corresponds to the search space of our stereo matching algorithm.

## 3 Pedestrian map computation

The proposed pedestrian detection method makes use of a Markov Random Field (MRF) based stereo matching. The objective is to minimize the pairwise MRF energy function:

$$E(l) = \sum_{p \in \mathcal{I}} D_p(l_p) + \lambda \sum_{(p,q) \in \mathcal{N}} V_{p,q}(l_p, l_q) \quad (2)$$

where: (i) $p$ is a pixel belonging to the image $\mathcal{I}$; (ii) given the label set $\mathcal{L}$, $l$ is a labeling assigning a value $l_p \in \mathcal{L}$ to each $p \in \mathcal{I}$; (iii) $\mathcal{N}$ is the set of edges of the image graph (4-connectivity is assumed); (iv) $D_p$ is the data cost function; (v) $V_{p,q}$ is the discontinuity cost function; (vi) $\lambda$ is a regularization parameter.

**Label definition: height-based optimization** A crucial point of our algorithm is the choice of the type of labels used. A common choice in state-of-the-art stereo matching is to use *depth* as label. Conversely, our method performs an optimization based on *height* labels. First, the choice of height is more natural if we consider how we build the search space (the volume between two planes at predefined heights). Moreover, the following task which benefits from a pedestrian detection map is the tracking on the reference plane, and, while height alone is enough to perform the ground projection (given the the estimated geometry), depth information needs to be converted nonlinearly into a ground-related variable (typically height) and regularization behaviors, particularly at discontinuities, are not equivalent following the different representations.

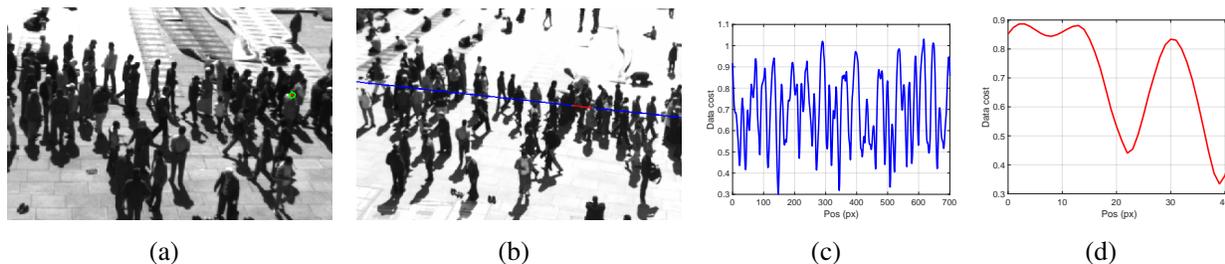

(a) (b) (c) (d)

Figure 3: DAISY dissimilarity: a) projected pixel b) search segment corresponding to $[h_{min}, h_{max}]$ c) DAISY dissimilarity along the entire epipolar line d) DAISY dissimilarity restricted to the search segment.

The height label allows us to define more sophisticated constraints on local image patches (e.g. head patches) without the need of a higher-order MRF. While constant depth assumption expresses heads as planes fronto-parallel to the camera plane, the height and vertical vanishing points can be exploited to constrain the head to resemble locally a plane perpendicular to the ground.

The label set $\mathcal{L} = \{h_{min}, h_{min} + \Delta_h, \ldots, h_{max}, u\}$ is defined in the interval $[h_{min}, h_{max}]$ with a sampling step $\Delta_h$, and is augmented by an *unknown* label $u$, meaning that no pedestrian is found at the specified location. We set $h_{min} = 140cm$, $h_{max} = 200cm$, $\Delta_h = 2.5cm$.

**Data cost function** The choice of a local region descriptor for dense matching is guided by the fact that our method is supposed to work even in a wide-baseline scenario with consistent perspective distortion. We employ the DAISY descriptor from Tola *et al.* [25], which has proven to be robust to perspective and illumination changes while showing a good computational efficiency. We express the data cost between $C_i$ and $C_j$ as the DAISY dissimilarity [25]:

$$D_p^{i,j}(l_p) = \frac{1}{S} \sum_{k=1}^{S} \| D_i^{[k]}(p) - D_j^{[k]}(\mathbf{H}_{i,j}^{l_p} p) \| \tag{3}$$

where $S$ is the number of histograms of the DAISY descriptors, $D_i^{[k]}(p)$ is the *k-th* histogram evaluated at pixel $p$ of camera $C_i$, $D_j^{[k]}(\mathbf{H}_{i,j}^{l_p} p)$ is the *k-th* histogram of the DAISY descriptor evaluated at the projection of pixel $p$ at an height $l_p$ on the *epipolar segment* at camera $C_j$, by using the variable-height homography $\mathbf{H}_{i,j}^{l_p}$. Figure 3 presents the typical behaviour of the dissimilarity, and how restricting accurately the search space to the epipolar segment helps.

The special *unknown* label $u$ is assigned with a constant data cost value $K_{d,u}$. The total data cost function $D_p(l_p)$ will be a combination of each $D_p^{i,j}(l_p)$ computed between the reference camera $C_i$ and any $C_j \in \mathcal{N}(C_i)$. Our experiments have been carried with $|\mathcal{N}| = 2$ neighbors, so a simple average of the two curves has demonstrated an effective combination, while with an increasing number of cameras an outlier robust cost merging method, like the one proposed by Vogiatzis *et al.* [27], is necessary.

**Discontinuity cost function** By using the heights as labels we need an efficient method to estimate at each pixel location the expected local height variation. Given a point $p \in \mathcal{I}$, the direction of maximum variation of the height will be along the line $\mathbf{r}_p$ connecting $p$ with the vertical vanishing point. In order to provide a fast and reliable estimation of the height variation around $p$, we consider a small patch area, with the length of an average head $L$, centered in $p$ as a planar surface. The direction of maximum height variation favors an orientation of the corresponding 3D patch which is perpendicular to the ground plane. We define as $|\nabla_p|$ the estimated absolute value of the height variation along $\mathbf{r}_p$ at a unit distance from $p$. Given the planar locality assumption, this quantity is expressed as:

$$|\nabla_p| = \frac{L}{d(p, p^L)} \quad (4)$$

where $p^L$ is the image projection of pixel $p$ into the 3D parallel plane located at distance $L$ from the 3D plane determined by $p$. The point $p^L$ is evaluated using homologies as: $p^L = \mathbf{B}^{\bar{h}-L}\left(\mathbf{B}^{\bar{h}}\right)^{-1} p$, where $\bar{h}$ is the central value of the height interval $[h_{min}, h_{max}]$. The choice of a constant $\bar{h}$ value is justified by the negligibly small variation of $|\nabla_p|$ for different $h$ values with respect to $\Delta_h$ and for a given $p$, leading to no effect on the final evaluation of the discontinuity function. As a consequence, Eq. (4) allows us to estimate the $|\nabla_p|$ map once during the algorithm initialization step.

Let us consider the neighbor points $p = (x_p, y_p), q = (x_q, y_q) \in \mathcal{I}$. The expected height variation between $p$ and $q$ is proportional to $|\nabla_p|$ and to the projection of $q$ on the line $\mathbf{r}_p$. The point $q^\perp = (x_{q^\perp}, y_{q^\perp})$ represents the orthogonal projection of $q$ on $\mathbf{r}_p$. In order for this projection to be valid, we use the hypothesis that for neighboring pixel in the image space, the angles with the $u$ axis of the $u$-$v$ image reference system of the lines $\mathbf{r}_p$ and $\mathbf{r}_q$ are almost identical: $\theta_{r_p} \approx \theta_{r_q}$. We can define the following distance function:

$$D_{p,q}(l_p, l_q) = \left| l_p - l_q - s_p |\nabla_p| d(p, q^\perp) \right| \quad (5)$$

where $s_p = 1$ if $y_p < y_{q^\perp}$ and $s_p = -1$ otherwise, meaning that the height value has to decrease when moving to lower pixels in the image space. Please note that since $\theta_{r_p} \approx \theta_{r_q}$ and $|\nabla_p| \approx |\nabla_q|$, it follows that $D_{p,q}(l_p, l_q) \approx D_{q,p}(l_q, l_p)$, but the equality does not hold strictly numerically (the maximum difference observed during experiments is of the order of $10^{-3}$). We enforce a symmetrical message flow by evaluating for each pair $p, q$ the distance functions in both directions, and by considering the one which provides the highest error. The discontinuity function between two labels which are not both unknown is defined as a truncated distance:

$$\hat{V}_{p,q}(l_p, l_q) = \min\left[\frac{\max\left(D_{p,q}(l_p, l_q), D_{q,p}(l_q, l_p)\right)}{\Delta_h}, K\right] \quad (6)$$

The total discontinuity cost function is expressed as:

$$V_{p,q}(l_p, l_q) = \begin{cases} \hat{V}_{p,q}(l_p, l_q) & l_p \neq u \wedge l_q \neq u \\ K_{V,u} & (l_p = u \wedge l_q \neq u) \vee (l_p \neq u \wedge l_q = u) \\ 0 & l_p = u \wedge l_q = u \end{cases} \quad (7)$$

where $K_{V,u}$ is a constant discontinuity cost for *unknown* labels (in the experiments $K_{V,u} = K$ for convenience).

The proposed pairwise discontinuity cost does not satisfy the submodularity property, therefore an alpha-expansion graph cut algorithm is not applicable to the optimization process. We provide results of the MRF optimization with Loopy Belief Propagation [28], but other techniques such as tree-reweighted message passing [15, 24] may be used as well.
**Temporal filtering** In order to illustrate qualitatively and quantitatively the interest of our work, we perform a simple temporal filtering of the detection results. The temporal filtering step outputs motion consistent trajectory fragments, denoted as tracklets. A more involved approach for the validation of instantaneous detections would require the use of appearance information either in the optimization or in a subsequent tracking algorithm, but these extensions go beyond the scope of the present purely geometry based work.

| $\lambda$ value | Recall | Precision |
|---|---|---|
| 0.08 | 89.89% | 42.65% |
| 0.15 | 74.37% | 66.78% |
| 0.20 | 62.23% | 82.82% |

Table 1: Recall and precision on the *Sparse* sequence depending on the regularization parameter $\lambda$. Here $\theta_l = 1$.

| $\theta_l$ value | Recall | Precision |
|---|---|---|
| 0 | 85.60% | 65.67% |
| 1 | 80.20% | 72.35% |
| 2 | 75.10% | 77.18% |
| 3 | 70.26% | 79.81% |

Table 2: Recall and precision on the *Dense* sequence depending on the tracklet threshold $\theta_l$.

The temporal filtering is applied as follows:
**1** The height map points are projected on the reference plane.
**2** Local maxima (in terms of height) are identified in the projected data.
**3** Each point is clustered around the closest local maximum, and a centroid is computed for each cluster.
**4** Tracklet creation and extension: each centroid may either be associated to an existing tracklet if it is located closer than $\theta_d$ from the linearly predicted tracklet location, and if its height is closer than $\theta_h$ to the tracklet average height; otherwise, a new tracklet is created. For all our experiments we used $\theta_d = 20cm$ and $\theta_h = 15cm$.
**5** Any tracklet which is not extended is terminated.

At the end of the temporal filtering part, tracklets of length equal or less than a tracklet thresholding parameter $\theta_l$ are discarded. For our experiments we considered values of $\theta_l \in \{0,\ldots,3\}$, with 0 meaning no filtering.

## 4  Experiments

We evaluate the proposed algorithm on a crowded scene dataset used initially by Pellicanò *et al.* [19] for wide baseline relative pose estimation. The dataset contains three synchronized video streams of a homogeneous crowd leaving the Regents Park Mosque in London. The laser measured distances between the central camera and the other two are 9.35 and 10.1 *m*. We present the results obtained in the central part of the scene which corresponds to the overlapping area of the three cameras, and which has roughly 400 $m^2$. For clarity, we recall that the related works on unsupervised detection of Eshel and Moses [6] and Khan and Shah [13] are not suitable for comparison in such kind of scenes. The method in [6] is based on scene constraints which are not transposable in a wide baseline open environment; the work in [13] tracks feet locations, and this operation is not applicable in higher density scenes as the one proposed.

In order to highlight better the specific behavior of our method, we process two different manually annotated sequences, the first one denoted as *Sparse* containing 200 frames and 2969 manually annotated pedestrians, and a second sequence denoted as *Dense* containing 500 frames and 18567 annotated pedestrians, most of them being clustered in a transit zone (Fig. 4). The ground truth annotations are used to evaluate the overall object level precision and the recall of our method in each sequence.

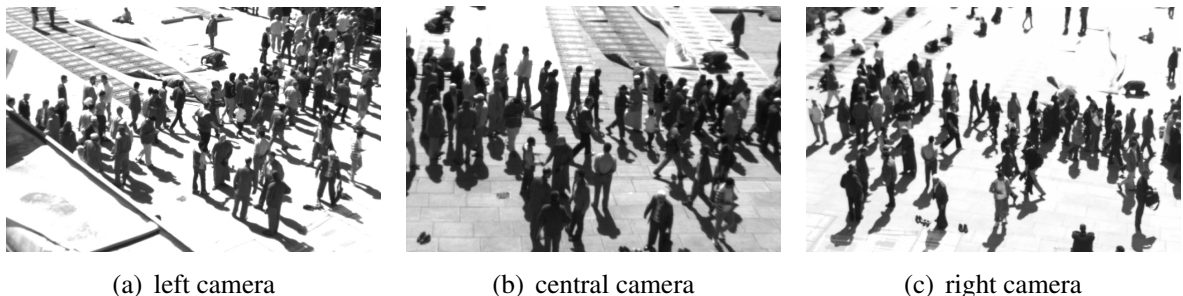

(a) left camera     (b) central camera     (c) right camera

Figure 4: Regents Park Mosque dataset, *Dense* sequence.

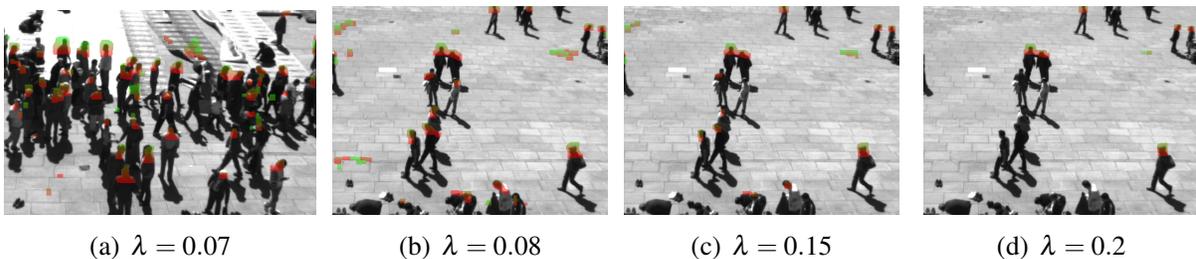

(a) $\lambda = 0.07$     (b) $\lambda = 0.08$     (c) $\lambda = 0.15$     (d) $\lambda = 0.2$

Figure 5: Detections on the (a) *Dense* sequence and on (b-d) *Sparse* sequence with varying levels of regularization. The tracklet threshold is set to $\theta_l = 1$.

**Impact of the regularization parameter** $\lambda$. The parameter $\lambda$ of Eq. (2) has a significant impact on the results as it controls the enforcement of the height gradient constraint with respect to the photometric term. In Table (1) we present the results in terms of precision and recall obtained on the *Sparse* sequence, using different values of $\lambda$ which cover the entire effective range. The tracklet threshold is set to $\theta_l = 1$. The impact of $\lambda$ is even more marked in a sparse setting, as the geometry constraint allows for removing the false positives created by the ground surface, in the absence of an object appearance model. Figures 5(b)-5(d) present the detection result on frame 27, with increasing values of $\lambda$.

**Impact of the tracklet threshold** $\theta_l$. The parameter $\theta_l$ controls in a simple manner the geometric consistency of the tracklet and is quite effective in removing spurious trajectories, which are uncommon in dense crowds. In Table (2) we present the results in terms of precision and recall obtained on the *Dense* sequence, using values of $\theta_l \in \{0, \ldots, 3\}$. The regularization parameter is set to $\lambda = 0.07$. Beyond these lengths, the majority of tracklets are correct and the thresholding becomes detrimental. Figure 5(a) presents the result of the detection on frame 5 of the *Dense* sequence.

**Final discussion of results and failure cases** The algorithm we propose clearly shows an excellent performance on strongly occluded crowd images, despite the use of only three views and the absence of appearance related terms. Besides, the influence of the parameters $\theta_l$ and $\lambda$ on the results may be exploited in order to favor precision or recall. Finally, due to the absence of appearance information in our framework, some expected difficulties arise. At low densities, ground areas may be visible and any ground related phantom will be persistent, and impossible to differentiate from a still individual (Fig. 5(b)) - we highlight this behaviour in Table 1, while it also worth noting that even in *Dense* the precision in Table 2 is impacted by some occasional frames with lower densities. A second failure case arises whenever in a detection blob corresponding to multiple pedestrians, there is only one local height maximum. In some cases, objects which are not heads and which are raised at above-torso level are correctly detected as being located at a valid height from the ground. We expect these cases to be easily dealt with by taking into account an appearance based model. Since we address the detection problem as an energy minimization, one straightforward way to extend our work is to include an additional term related to appearance. Alternatively, a standard tracking algorithm including appearance cues would address these cases as well.

## 5   Conclusion

We proposed a method for locating pedestrian heads in a cluttered outdoor scene using a multiple camera network under realistic conditions of size, illumination and pose variations. With this advancement, our principal objective is to provide a detection algorithm based entirely on geometric constraints, which may be complemented using different approaches with appearance based terms. Alternatively, the result of our algorithm may be included as a detection prior in a detection or tracking algorithm relying mainly on visual cues.

**Acknowledgment** This work was supported by ANR grant ANR-15-CE39-0005 and by QNRF grant NPRP-09-768-1-114. We gratefully acknowledge the support from Regent's Park Mosque for providing access to the site during data collection.

## References

[1] Ákos Kiss and Tamás Szirányi. Localizing people in multi-view environment using height map reconstruction in real-time. *Pattern Recognition Letters*, 34(16):2135 – 2143, 2013. ISSN 0167-8655.

[2] Alexandre Alahi, Laurent Jacques, Yannick Boursier, and Pierre Vandergheynst. Sparsity driven people localization with a heterogeneous network of cameras. *Journal of Mathematical Imaging and Vision*, 41(1-2):39–58, 2011.

[3] Ming-Ching Chang, Nils Krahnstoever, and Weina Ge. Probabilistic group-level motion analysis and scenario recognition. In *Computer Vision (ICCV), 2011 IEEE International Conference on*, pages 747–754. IEEE, 2011.

[4] Antonio Criminisi, Ian Reid, and Andrew Zisserman. Single view metrology. In *Computer Vision, 1999. The Proceedings of the Seventh IEEE International Conference on*, volume 1, pages 434–441. IEEE, 1999.

[5] Dima Damen and David Hogg. Detecting carried objects from sequences of walking pedestrians. *IEEE transactions on pattern analysis and machine intelligence*, 34(6): 1056–1067, 2012.

[6] Ran Eshel and Yael Moses. Tracking in a dense crowd using multiple cameras. *International Journal of Computer Vision*, 88(1):129–143, 2010. ISSN 09205691. doi: 10.1007/s11263-009-0307-0.

[7] F. Fleuret, J. Berclaz, R. Lengagne, and P. Fua. Multi-camera people tracking with a probabilistic occupancy map. *IEEE Transactions on Pattern Analysis and Machine Intelligence (TPAMI)*, 30(2):267–282, 2008.

[8] Weina Ge and Robert T Collins. Marked point processes for crowd counting. In *Computer Vision and Pattern Recognition, 2009. CVPR 2009. IEEE Conference on*, pages 2913–2920. IEEE, 2009.

[9] Israel Dejene Gebru, Xavier Alameda-Pineda, Florence Forbes, and Radu Horaud. Em algorithms for weighted-data clustering with application to audio-visual scene analysis. *IEEE Trans. on pattern analysis and machine intelligence*, 38(12):2402–2415, 2016.

[10] Li Guan, Jean-Sébastien Franco, and Marc Pollefeys. Multi-view occlusion reasoning for probabilistic silhouette-based dynamic scene reconstruction. *International journal of computer vision*, 90(3):283–303, 2010.

[11] Raffay Hamid, Ramkrishan Kumar, Jessica Hodgins, and Irfan Essa. A visualization framework for team sports captured using multiple static cameras. *Computer Vision and Image Understanding*, 118:171–183, 2014.

[12] Zhixing Jin, Le An, and Bir Bhanu. Group structure preserving pedestrian tracking in a multi-camera video network. *IEEE Transactions on Circuits and Systems for Video Technology*, 2016.

[13] Saad M Khan and Mubarak Shah. Tracking multiple occluding people by localizing on multiple scene planes. *IEEE transactions on pattern analysis and machine intelligence*, 31(3):505–519, 2009.

[14] Kyungnam Kim and Larry S Davis. Multi-camera tracking and segmentation of occluded people on ground plane using search-guided particle filtering. In *European Conference on Computer Vision*, pages 98–109. Springer, 2006.

[15] Vladimir Kolmogorov. Convergent tree-reweighted message passing for energy minimization. *IEEE transactions on pattern analysis and machine intelligence*, 28(10):1568–1583, 2006.

[16] José Lezama, Rafael Grompone von Gioi, Gregory Randall, and Jean-Michel Morel. Finding vanishing points via point alignments in image primal and dual domains. In *Proceedings of the IEEE Conference on Computer Vision and Pattern Recognition*, pages 509–515, 2014.

[17] Martijn C Liem and Dariu M Gavrila. Joint multi-person detection and tracking from overlapping cameras. *Computer Vision and Image Understanding*, 128:36–50, 2014.

[18] Anurag Mittal and Larry S Davis. M2tracker: A multi-view approach to segmenting and tracking people in a cluttered scene. *Int. Journal of Computer Vision*, 51(3):189–203, 2003.

[19] Nicola Pellicanò, Emanuel Aldea, and Sylvie Le Hégarat-Mascle. Robust wide baseline pose estimation from video. In *Proceedings of the International Conference on Pattern Recognition (ICPR)*, 2016.

[20] Peixi Peng, Yonghong Tian, Yaowei Wang, Jia Li, and Tiejun Huang. Robust multiple cameras pedestrian detection with multi-view bayesian network. *Pattern Recognition*, 48(5):1760–1772, 2015.

[21] Horst Possegger, Sabine Sternig, Thomas Mauthner, Peter M Roth, and Horst Bischof. Robust real-time tracking of multiple objects by volumetric mass densities. In *Proceedings of the IEEE Conference on Computer Vision and Pattern Recognition*, pages 2395–2402, 2013.

[22] Taiki Sekii. Robust, real-time 3d tracking of multiple objects with similar appearances. In *Proceedings of the IEEE Conference on Computer Vision and Pattern Recognition*, pages 4275–4283, 2016.

[23] Sabine Sternig, Thomas Mauthner, Arnold Irschara, Peter M Roth, and Horst Bischof. Multi-camera multi-object tracking by robust hough-based homography projections. In *Computer Vision Workshops (ICCV Workshops), 2011 IEEE International Conference on*, pages 1689–1696. IEEE, 2011.

[24] Richard Szeliski, Ramin Zabih, Daniel Scharstein, Olga Veksler, Vladimir Kolmogorov, Aseem Agarwala, Marshall Tappen, and Carsten Rother. A comparative study of energy minimization methods for markov random fields with smoothness-based priors. *IEEE transactions on pattern analysis and machine intelligence*, 30(6): 1068–1080, 2008.

[25] Engin Tola, Vincent Lepetit, and Pascal Fua. Daisy: An efficient dense descriptor applied to wide-baseline stereo. *IEEE transactions on pattern analysis and machine intelligence*, 32(5):815–830, 2010.

[26] Akos Utasi and Csaba Benedek. A 3-d marked point process model for multi-view people detection. In *Computer Vision and Pattern Recognition (CVPR), 2011 IEEE Conference on*, pages 3385–3392. IEEE, 2011.

[27] George Vogiatzis, Carlos Hernández Esteban, Philip HS Torr, and Roberto Cipolla. Multiview stereo via volumetric graph-cuts and occlusion robust photo-consistency. *IEEE Transactions on Pattern Analysis and Machine Intelligence*, 29(12):2241–2246, 2007.

[28] Jonathan S Yedidia, William T Freeman, Yair Weiss, et al. Generalized belief propagation. In *NIPS*, volume 13, pages 689–695, 2000.